\documentclass[11pt,a4paper]{article}
\usepackage[hyperref]{naaclhlt2018}
\usepackage{times}
\usepackage{latexsym}
\usepackage{amsmath}
\usepackage{graphicx}
\usepackage{booktabs}
\usepackage{multirow}
\usepackage{comment}
\usepackage{url}
\usepackage{algorithm,algorithmicx,algpseudocode}
\makeatletter
\def\BState{\State\hskip-\ALG@thistlm}
\makeatother

\newcommand{\expect}[1]{\underset{#1}{E}}

\aclfinalcopy 

\title{Variational Knowledge Graph Reasoning}

\author{Wenhu Chen, Wenhan Xiong, Xifeng Yan, William Yang Wang\\
  Department of Computer Science\\
  University of California, Santa Barbara\\
  Santa Barbara, CA 93106 \\
  {\tt \{wenhuchen,xwhan,xyan,william\}@cs.ucsb.edu} \\}

\date{}

\begin{document}
\maketitle
\begin{abstract}
Inferring missing links in knowledge graphs (KG) has attracted a lot of attention from the research community. In this paper, we tackle a practical query answering task involving predicting the relation of a given entity pair. We frame this prediction problem as an inference problem in a probabilistic graphical model and aim at resolving it from a variational inference perspective. In order to model the relation between the query entity pair, we assume that there exists an underlying latent variable (paths connecting two nodes) in the KG, which carries the equivalent semantics of their relations. However, due to the intractability of connections in large KGs, we propose to use variation inference to maximize the evidence lower bound. More specifically, our framework (\textsc{Diva}) is composed of three modules, i.e. a posterior approximator, a prior (path finder), and a likelihood (path reasoner). By using variational inference, we are able to incorporate them closely into a unified architecture and jointly optimize them to perform KG reasoning. With active interactions among these sub-modules, \textsc{Diva} is better at handling noise and coping with more complex reasoning scenarios. In order to evaluate our method, we conduct the experiment of the link prediction task on multiple datasets and achieve state-of-the-art performances on both datasets.
\end{abstract}

\section{Introduction}
Large-scaled knowledge graph supports a lot of downstream natural language processing tasks like question answering, response generation, etc. However, there are large amount of important facts missing in existing KG, which has significantly limited the capability of KG's application. Therefore, automated reasoning, or the ability for computing systems to make new inferences from the observed evidence, has attracted lots of attention from the research community. In recent years, there are surging interests in designing machine learning algorithms for complex reasoning tasks, especially in large knowledge graphs (KGs) where the countless entities and links have posed great challenges to traditional logic-based algorithms. Specifically, we situate our study in this large KG multi-hop reasoning scenario, where the goal is to design an automated inference model to complete the missing links between existing entities in large KGs. For examples, if the KG contains a fact like \textit{president}(\textit{BarackObama}, \textit{USA}) and \textit{spouse}(\textit{Michelle, BarackObama}), then we would like the machines to complete the missing link \textit{livesIn}(\textit{Michelle}, \textit{USA}) automatically. Systems for this task are essential to complex question answering applications.  

To tackle the multi-hop link prediction problem, various approaches have been proposed. Some earlier works like PRA~\cite{lao2011random,gardner2014incorporating,gardner2013improving} use bounded-depth random walk with restarts
to obtain paths. More recently, DeepPath~\cite{xiong2017deeppath} and MINERVA~\cite{das2017go}, frame the path-finding problem as a Markov Decision Process (MDP) and utilize reinforcement learning (RL) to maximize the expected return. Another line of work along with ours are Chain-of-Reasoning~\cite{das2016chains} and Compositional Reasoning~\cite{neelakantan2015compositional}, which take multi-hop chains learned by PRA as input and aim to infer its relation.

Here we frame the KG reasoning task as a two sub-steps, i.e. ``Path-Finding'' and ``Path-Reasoning''. We found that most of the related research is only focused on one step, which leads to major drawbacks---lack of interactions between these two steps. More specifically, DeepPath~\cite{xiong2017deeppath} and MINERVA~\cite{das2017go} can be interpreted as enhancing the ``Path-Finding'' step while compositional reasoning~\cite{neelakantan2015compositional} and chains of reasoning~\cite{das2016chains} can be interpreted as enhancing the ``Path-Reasoning'' step. DeepPath is trained to find paths more efficiently between two given entities while being agnostic to whether the entity pairs are positive or negative, whereas MINERVA learns to reach target nodes given an entity-query pair while being agnostic to the quality of the searched path\footnote{MINERVA assigns constant rewards to all paths reaching the destination while ignoring their qualities.}. In contrast, chains of reasoning and compositional reasoning only learn to predict relation given paths while being agnostic to the path-finding procedure. The lack of interaction prevents the model from understanding more diverse inputs and make the model very sensitive to noise and adversarial samples. 

In order to increase the robustness of existing KG reasoning model and handle noisier environments, we propose to combine these two steps together as a whole from the perspective of the latent variable graphic model. This graphic model views the paths as discrete latent variables and relation as the observed variables with a given entity pair as the condition, thus the path-finding module can be viewed as a prior distribution to infer the underlying links in the KG. In contrast, the path-reasoning module can be viewed as the likelihood distribution, which classifies underlying links into multiple classes. With this assumption, we introduce an approximate posterior and design a variational auto-encoder~\cite{kingma2013auto} algorithm to maximize the evidence lower-bound. This variational framework closely incorporates two modules into a unified framework and jointly train them together. By active cooperations and interactions, the path finder can take into account the value of searched path and resort to the more meaningful paths. Meanwhile, the path reasoner can receive more diverse paths from the path finder and generalizes better to unseen scenarios.
Our contributions are three-fold:
\begin{itemize}
\item We introduce a variational inference framework for KG reasoning, which tightly integrates the path-finding and path-reasoning processes to perform joint reasoning.
\item We have successfully leveraged negative samples into training and increase the robustness of existing KG reasoning model.
\item We show that our method can scale up to large KG and achieve state-of-the-art results on two popular datasets.
\end{itemize}

The rest of the paper is organized as follow. In Section~\ref{sec:related} we will outline related work on KG embedding, multi-hop reasoning, and variational auto-encoder. We describe our variational knowledge reasoner \textsc{Diva} in Section~\ref{sec:model}. Experimental results are presented in Section~\ref{sec:exp}, and we conclude in Section~\ref{sec:conclusion}.

\section{Related Work}
\label{sec:related}
\subsection{Knowledge Graph Embeddings}
Embedding methods to model multi-relation data from KGs have been extensively studied in recent years~\cite{nickel2011three,bordes2013translating,socher2013reasoning,lin2015learning,trouillon2017knowledge}. From a representation learning perspective, all these methods are trying to learn a projection from symbolic space to vector space. For each triple $(e_s, r, e_d)$ in the KG, various score functions can be defined using either vector or matrix operations. Although these embedding approaches have been successful capturing the semantics of KG symbols (entities and relations) and achieving impressive results on knowledge base completion tasks, most of them fail to model multi-hop relation paths, which are indispensable for more complex reasoning tasks. Besides, since all these models operate solely on latent space, their predictions are barely interpretable.

\subsection{Multi-Hop Reasoning}
The Path-Ranking Algorithm (PRA) method is the first approach to use a random walk with restart mechanism to perform multi-hop reasoning. Later on, some research studies~\cite{gardner2014incorporating,gardner2013improving} have revised the PRA algorithm to compute feature similarity in the vector space. These formula-based algorithms can create a large fan-out area, which potentially undermines the inference accuracy. To mitigate this problem, a Convolutional Neural Network(CNN)-based model~\cite{toutanova2015representing} has been proposed to perform multi-hop reasoning. Recently, DeepPath~\cite{xiong2017deeppath} and MINERVA~\cite{das2017go} view the multi-hop reasoning problem as a Markov Decision Process, and leverages REINFORCE~\cite{williams1992simple} to efficiently search for paths in large knowledge graph. These two methods are reported to achieve state-of-the-art results, however, these two models both use heuristic rewards to drive the policy search, which could make their models sensitive to noises and adversarial examples. 

\subsection{Variational Auto-encoder}
Variational Auto-Encoder~\cite{kingma2013auto} is a very popular algorithm to perform approximate posterior inference in large-scale scenarios, especially in neural networks. Recently, VAE has been successfully applied to various complex machine learning tasks like image generation~\cite{mansimov2015generating}, machine translation~\cite{zhang2016variational}, sentence generation~\cite{guu2017generating} and question answering~\cite{zhang2017variational}. ~\citet{zhang2017variational} is closest to ours, this paper proposes a variational framework to understand the variability of human language about entity referencing. In contrast, our model uses a variational framework to cope with the complex link connections in large KG. Unlike the previous research in VAE, both ~\citet{zhang2017variational} and our model both use discrete variable as the latent representation to infer the semantics of given entity pairs. More specifically, we view the generation of relation as a stochastic process controlled by a latent representation, i.e. the connected multi-hop link existed in the KG. Though the potential link paths are discrete and countable, its amount is still very large and poses challenges to direct optimization. Therefore, we resort to variational auto-encoder as our approximation strategy.

\section{Our Approach}
\label{sec:model}
\subsection{Background}
Here we formally define the background of our task. Let $\mathcal{E}$ be the set of entities and $R$ be the set of relations. Then a KG is defined as a collection of triple facts $(e_s, r, e_d)$, where $e_s, e_d \in \mathcal{E}$ and $r \in \mathcal{R}$. We are particularly interested in the problem of relation inference, which seeks to answer the question in the format of $(e_s, ?, e_d)$, the problem setting is slightly different from standard link prediction to answer the question of $(e_s, r, ?)$. Next, in order to tackle this classification problem, we assume that there is a latent representation for given entity pair in the KG, i.e. the collection of linked paths, these hidden variables can reveal the underlying semantics between these two entities. Therefore, the link classification problem can be decomposed into two modules -- acquire underlying paths (Path Finder) and infer relation from latent representation (Path Reasoner).
\paragraph{Path Finder}
The state-of-the-art approach~\cite{xiong2017deeppath,das2017go} is to view this process as a Markov Decision Process (MDP). A tuple $<S, A, P>$ is defined to represent the MDP, where $S$ denotes the current state, e.g. the current node in the knowledge graph, $A$ is the set of available actions, e.g. all the outgoing edges from the state, while $P$ is the transition probability describing the state transition mechanism. In the knowledge graph, the transition of the state is deterministic, so we do not need to model the state transition $P$.
\paragraph{Path Reasoner}
The common approach~\cite{lao2011random,neelakantan2015compositional,das2016chains} is to encode the path as a feature vector and use a multi-class discriminator to predict the unknown relation. PRA~\cite{lao2011random} proposes to encode paths as binary features to learn a log-linear classifier, while ~\cite{das2016chains} applies recurrent neural network to recursively encode the paths into hidden features and uses vector similarity for classification. 
\subsection{Variational KG Reasoner (\textsc{Diva})}
\begin{figure}[t]
\centering
\includegraphics[width=\linewidth]{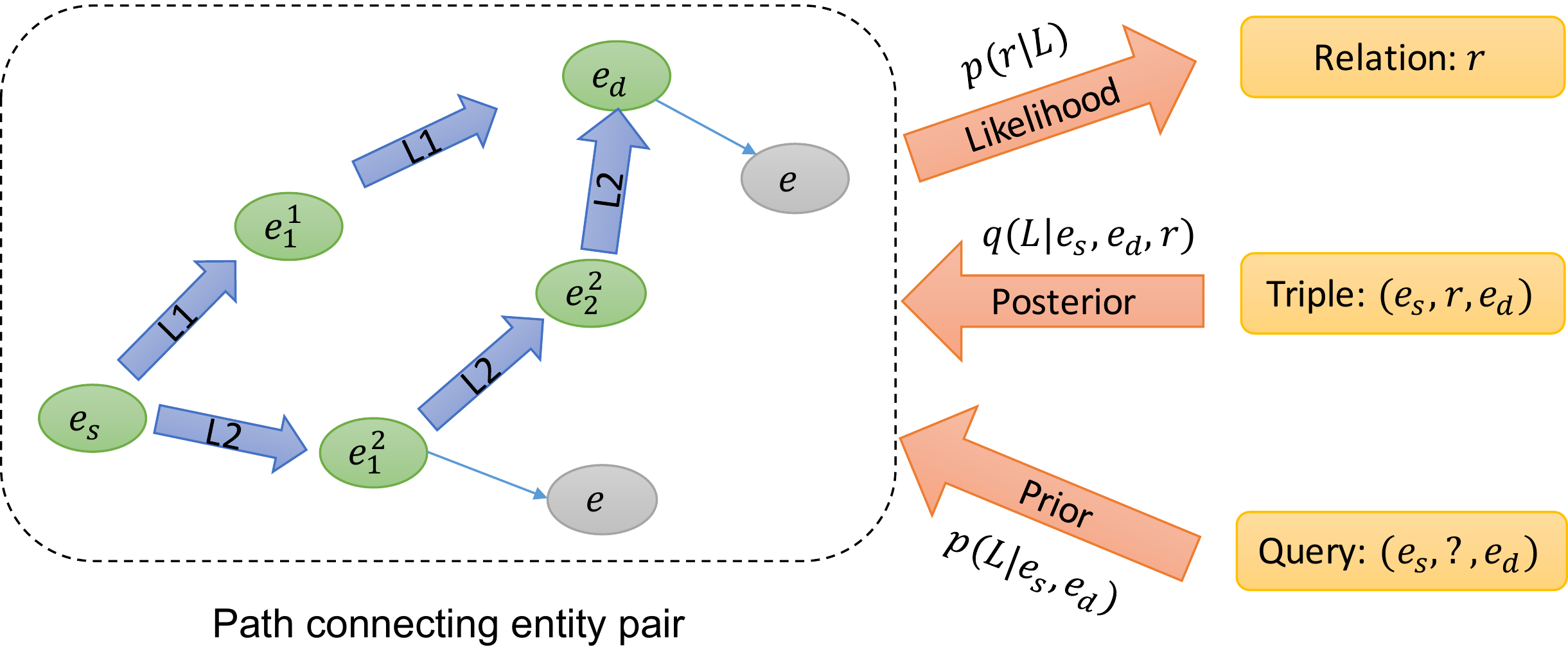}
\caption{The probabilistic graphical model of our proposed approach. Arrows with dotted border represent the approximate posterior, which is modeled as a multinomial distribution over the whole link space. Arrows with solid border represent the prior and likelihood distributions. }
\label{fig:graphic-model}
\end{figure}

\begin{figure}[t]
\centering
\includegraphics[width=\linewidth]{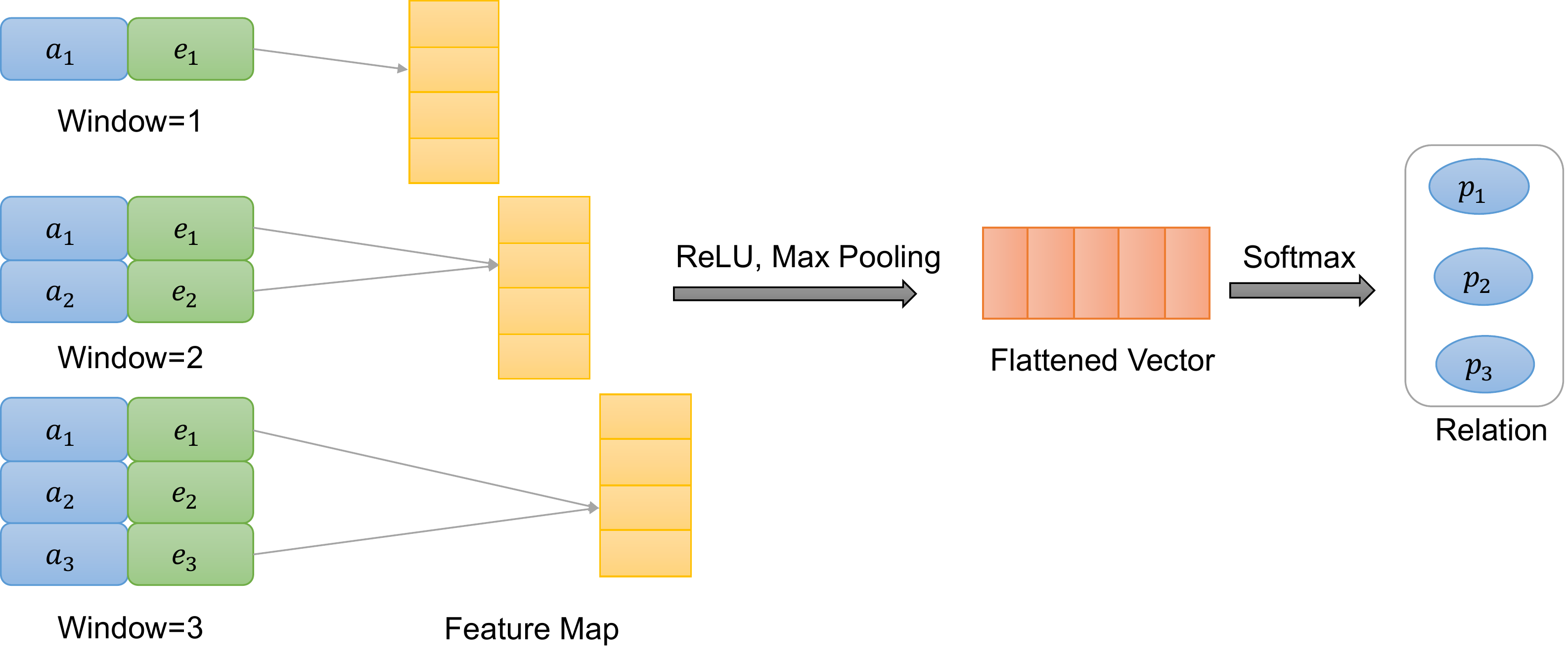}
\caption{Overview of the CNN Path Reasoner.}
\label{fig:conv}
\end{figure}

Here we draw a schematic diagram of our model in~\autoref{fig:graphic-model}. Formally, we define the objective function for the general relation classification problem as follows:
\begin{align}
\small
\begin{split}
Obj =& \sum_{(e_s, r, e_d) \in D} \log p(r|(e_s, e_d))\\
  =& \sum_{(e_s, r, e_d) \in D} \log \sum_L p_{\theta}(L|(e_s, e_d)) p(r|L)
\end{split}
\end{align}
where $D$ is the dataset, $(e_s, r, e_d)$ is the triple contained in the dataset, and $L$ is the latent connecting paths. The evidence probability $p(r|(e_s, e_d))$ can be written as the marginalization of the product of two terms over the latent space.
However, this evidence probability is intractable since it requires summing over the whole latent link space. Therefore, we propose to maximize its variational lower bound as follows:
\begin{align}
\small
\begin{split}
ELBO =& \expect{L \sim q_{\varphi}(L|r, (e_s, e_d))} [\log p_{\theta}(r|L)] - \\
         & D_{KL}(q_{\varphi}(L|r, (e_s, e_d))||p_{\beta}(L|(e_s, e_d)))
\end{split}
\end{align}
Specifically, the ELBO~\cite{kingma2013auto} is composed of three different terms -- likelihood $p_{\theta}\big(r|L)$, prior $p_{\beta}\big(L|(e_s, e_t))$, and posterior $q_{\varphi}\big(L|(e_s, e_d),r)$. In this paper, we use three neural network models to parameterize these terms and then follow~\cite{kingma2013auto} to apply variational auto-encoder to maximize the approximate lower bound. We describe these three models in details below:

\paragraph{Path Reasoner (Likelihood).} Here we propose a path reasoner using Convolutional Neural Networks (CNN)~\cite{lecun1995convolutional} and a feed-forward neural network. This model takes path sequence $L = \{a_1, e_1, \cdots, a_i, e_i, \cdots a_n, e_n\}$ to output a softmax probability over the relations set $R$, where $a_i$ denotes the $i$-th intermediate relation and $e_i$ denotes the $i$-th intermediate entity between the given entity pair. Here we first project them into embedding space and concatenate i-th relation embedding with $i$-th entity embedding as a combined vector, which we denote as $\{ f_1, f_2, \cdots, f_n \}$ and $f_i \in \mathcal{R}^{2E}$. As shown in~\autoref{fig:conv}, we pad the embedding sequence to a length of $N$. Then we design three convolution layers with window size of $(1 \times 2E), (2 \times 2E), (3 \times 2E)$, input channel size $1$ and filter size $D$. After the convolution layer, we use $(N \times 1), (N-1 \times 1), (N-2 \times 1)$ to max pool the convolution feature map. Finally, we concatenate the three vectors as a combined vector $F \in \mathcal{R}^{3D}$. Finally, we use two-layered MLP with intermediate hidden size of $M$ to output a softmax distribution over all the relations set $R$.
\begin{gather}
F = f(f_1, f_2, \cdots, f_N)\\
p(r|L; \theta) = softmax(W_r F + b_r)
\end{gather}
where $f$ denotes the convolution and max-pooling function applied to extract reasoning path feature $F$, and $W_r, b_r$ denote the weights and bias for the output feed-forward neural network.

\paragraph{Path Finder (Prior).} Here we formulate the path finder $p(L|(e_s, e_d))$ as an MDP problem, and recursively predict actions (an outgoing relation-entity edge $(a, e)$) in every time step based on the previous history $h_{t-1}$ as follows:
\begin{gather}
c_t = ReLU(W_{h} [h_t;e_d] + b_{h})\\
p((a_{t+1}, e_{t+1}) | h_t, \beta) = softmax(A_t c_t)
\end{gather}

\begin{figure}[t]
\centering
\includegraphics[width=1.0\linewidth]{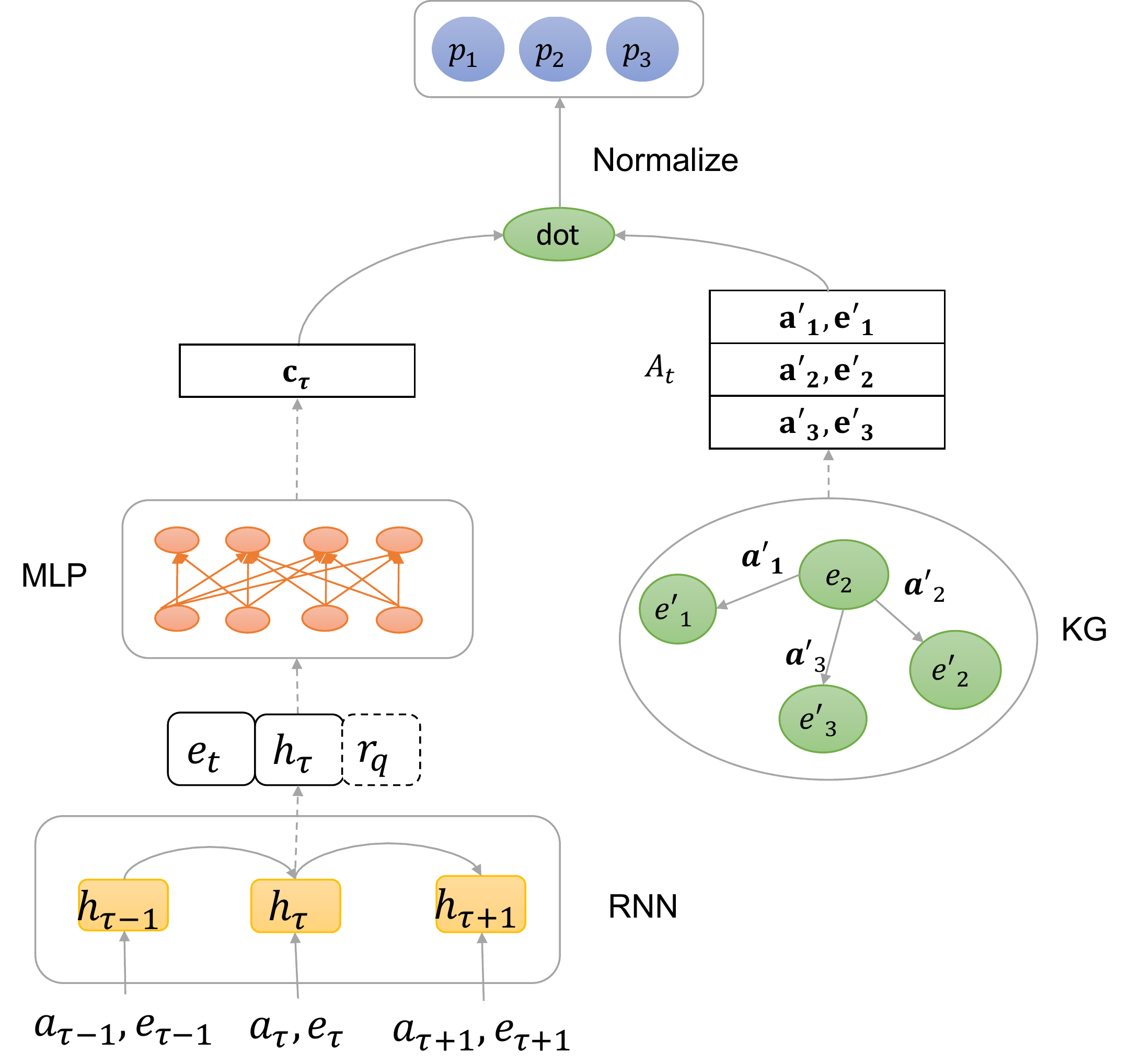}
\caption{An overview of the path finder model. Note that $r_q$ (query relation) exists in the approximate posterior while disappearing in the path finder model and $e_t$ represents the target entity embedding, $c_{\tau}$ is the output of MLP layer at time step $\tau$, $a', e'$ denotes the connected edges and ends in the knowledge graphs.}
\label{fig:rnn}
\end{figure}where the $h_t \in \mathcal{R}^H$ denotes the history embedding, $e_d \in \mathcal{R}^E$ denotes the entity embedding, $A_t \in \mathcal{R}^{|A| \times 2E}$ is outgoing matrix which stacks the concatenated embeddings of all outgoing edges and $|A|$ denotes the number of outgoing edge, we use $W_h$ and $b_h$ to represent the weight and bias of the feed-forward neural network outputting feature vector $c_t \in \mathcal{R}^{2E}$. The history embedding $h_t$ is obtained using an LSTM network~\cite{hochreiter1997long} to encode all the previous decisions as follows:
\begin{align}
h_t = LSTM(h_{t-1}, (a_t, e_t))
\end{align}
As shown in~\autoref{fig:rnn}, the LSTM-based path finder interacts with the KG in every time step and decides which outgoing edge $(a_{t+1}, e_{t+1})$ to follow, search procedure will terminate either the target node is reached or the maximum step is reached.

\paragraph{Approximate Posterior.} We formulate the posterior distribution $q(L|(e_s, e_d), r)$ following the similar architecture as the prior. The main difference lies in the fact that posterior approximator is aware of the relation $r$, therefore making more relevant decisions. The posterior borrows the history vector from finder as $h_t$, while the feed-forward neural network is distinctive in that it takes the relation embedding also into account. Formally, we write its outgoing distribution as follows:
\begin{align}
\begin{split}
u_t = ReLU(W_{hp} [h_t;e_d;r] + b_{hp})\\
q((a_{t+1}, e_{t+1}) | h_t; \varphi) = softmax(A_t u_t)
\end{split}
\end{align}
where $W_{hp}$ and $b_{hp}$ denote the weight and bias for the feed-forward neural network.

\subsection{Optimization}
In order to maximize the ELBO with respect to the neural network models described above, we follow VAE~\cite{kingma2013auto} to interpret the negative ELBO as two separate losses and minimize these them jointly using a gradient descent: 
\paragraph{Reconstruction Loss.} Here we name the first term of negative ELBO as reconstruction loss:
\begin{align}
J_R = \expect{L \sim q_{\varphi}(L|r, (e_s, e_d))} [-\log p_{\theta}(r|L)]
\end{align}
this loss function is motivated to reconstruct the relation $R$ from the latent variable $L$ sampled from approximate posterior, optimizing this loss function jointly can not only help the approximate posterior to obtain paths unique to particular relation $r$, but also teaches the path reasoner to reason over multiple hops and predict the correct relation.
\paragraph{KL-divergence Loss.} We name the second term as KL-divergence loss:
\begin{align}
\small
J_{KL} = D_{KL}(q_{\varphi}(L|r, e_s, e_d)|p_{\beta}(L|e_s, e_d))
\end{align}
this loss function is motivated to push the prior distribution towards the posterior distribution. The intuition of this loss lies in the fact that an entity pair already implies their relation, therefore, we can teach the path finder to approach the approximate posterior as much as possible. During test-time when we have no knowledge about relation, we use path finder to replace posterior approximator to search for high-quality paths.
\paragraph{Derivatives.} We show the derivatives of the loss function with respect to three different models. For the approximate posterior, we re-weight the KL-diverge loss and design a joint loss function as follows:
\begin{align}
J = J_R + \lambda_{KL} J_{KL}
\end{align}
where $\lambda_{KL}$ is the re-weight factor to combine these two losses functions together. Formally, we write the derivative of posterior as follows:
\begin{align}
\small
\begin{split}
\frac{\partial J}{\partial \varphi} =\expect{L \sim q_{\varphi}(L))}[& -f_{re}(L) \frac{\partial \log{q_{\varphi}(L| (e_s, e_d), r)}}{\partial \varphi}]
\end{split}
\label{eq:path-finder}
\end{align}
where $f_{re}(L) = \log{p_{\theta}} + \lambda_{KL} \log\frac{p_{\beta}}{q_{\varphi}}$ denotes the probability assigned by path reasoner. In practice, we found that a large KL-regularizer $\log \frac{p_{\beta}}{q_{\varphi}}$ causes severe instability during training, therefore we keep a low $\lambda_{KL}$ value during our experiments~\footnote{we set $\lambda_{KL}=0$ through our experiments.}.
For the path reasoner, we also optimize its parameters $\theta$ with regard to the reconstruction as follows:
\begin{align}
\frac{\partial J_R}{\partial \theta} = \expect{L \sim q_{\varphi}(L)} -\frac{\partial \log{p_{\theta}(r|L)}}{\partial \theta}
\label{eq:path-reasoner}
\end{align}
For the path finder, we optimize its parameters $\beta$ with regard to the KL-divergence to teach it to infuse the relation information into the found links.
\begin{align}
\frac{\partial J_{KL}}{\partial \beta} = \expect{L \sim q_{\varphi}(L)} -\frac{\partial \log{p_{\beta}(L|(e_s, e_d))}}{\partial \beta}
\label{eq:prior}
\end{align}

\begin{algorithm}[t]
\caption{The \textsc{Diva} Algorithm.}\label{alg:diva}
\begin{algorithmic}[1]
\Procedure{Training \& Testing}{}
\BState \emph{Train}:
\For{episode $\leftarrow$ 1 to N}
\State Rollout K paths from posterior $p_{\varphi}$
\If{Train-Posterior}
\State $\varphi \leftarrow \varphi - \eta \times \frac{\partial L_r}{\partial \varphi}$ 
\ElsIf{Train-Likelihood}
\State $\theta \leftarrow \theta - \eta  \times \frac{\partial L_r}{\partial \theta}$ 
\ElsIf{Train-Prior}
\State $\beta \leftarrow \beta - \eta  \times \frac{\partial L_{KL}}{\partial \beta}$ 
\EndIf
\EndFor
\BState \emph{Test MAP}:
\State Restore initial parameters $\theta, \beta$
\State Given sample $(e_s, r_q, (e_1, e_2, \cdots, e_n))$
\State $L_i \leftarrow BeamSearch(p_{\beta}(L|e_s, e_i))$
\State $S_i \leftarrow \frac{1}{|L_i|}\sum_{l \in L_i} p_{\theta}(r_q|l)$
\State Sort $S_i$ and find positive rank $ra^+$
\State $MAP \leftarrow \frac{1}{1 + ra^+}$
\EndProcedure
\end{algorithmic}
\end{algorithm}
\paragraph{Train \& Test} During training time, in contrast to the preceding methods like~\newcite{das2017go,xiong2017deeppath}, we also exploit negative samples by introducing an pseudo ``n/a'' relation, which indicates ``no-relation'' between two entities. Therefore, we manage to decompose the data sample $(e_q, r_q, [e_1^-, e_2^-, \cdots, e_n^+])$ into a series of tuples $(e_q, r_q', e_i)$, where $r_q'=r_q$ for positive samples and $r_q'=n/a$ for negative samples. During training, we alternatively update three sub-modules with SGD. During test, we apply the path-finder to beam-search the top paths for all tuples and rank them based on the scores assign by path-reasoner. More specifically, we demonstrate the pseudo code in~\autoref{alg:diva}.

\subsection{Discussion}
We here interpret the update of the posterior approximator in equation~\autoref{eq:path-finder} as a special case of REINFORCE~\cite{williams1992simple}, where we use Monte-Carlo sampling to estimate the expected return $\log p_{\theta}(r|L)$ for current posterior policy. This formula is very similar to DeepPath and MINERVA~\cite{xiong2017deeppath,das2017go} in the sense that path-finding process is described as an exploration process to maximize the policy's long-term reward. Unlike these two models assigning heuristic rewards to the policy, our model assigns model-based reward $\log{p_{\theta}(r|L)}$, which is known to be more sophisticated and considers more implicit factors to distinguish between good and bad paths.
Besides, our update formula for path reasoner~\autoref{eq:path-reasoner} is also similar to chain-of-reasoning~\cite{das2016chains}, both models are aimed at maximizing the likelihood of relation given the multi-hop chain. However, our model is distinctive from theirs in a sense that the obtained paths are sampled from a dynamic policy, by exposing more diverse paths to the path reasoner, it can generalize to more conditions. By the active interactions and collaborations of two models, \textsc{Diva} is able to comprehend more complex inference scenarios and handle more noisy environments.    

\section{Experiments}
\label{sec:exp}
To evaluate the performance of \textsc{Diva}, we explore the standard link prediction task on two different-sized KG datasets and compare with the state-of-the-art algorithms. Link prediction is to rank a list of target entities $(e_1^-, e_2^-, \cdots, e_n^+)$ given a query entity $e_q$ and query relation $r_q$. The dataset is arranged in the format of $(e_q, r_q, [e_1^-, e_2^-, \cdots, e_n^+])$, and the evaluation score (Mean Averaged Precision, MAP) is based on the ranked position of the positive sample.

\subsection{Dataset and Setting}
We perform experiments on two datasets, and the details of the statistics are described in~\autoref{tab:stat}. The samples of FB15k-237~\cite{toutanova2015representing} are sampled from FB15k~\cite{bordes2013translating}, here we follow DeepPath~\cite{xiong2017deeppath} to select 20 relations including Sports, Locations, Film, etc. Our NELL dataset is downloaded from the released dataset\footnote{https://github.com/xwhan/DeepPath}, which contains 12 relations for evaluation. Besides, both datasets contain negative samples obtained by using the PRA code released by~\newcite{lao2011random}. For each query $r_q$, we remove all the triples with $r_q$ and $r_q^{-1}$ during reasoning. During training, we set number of rollouts to 20 for each training sample and update the posterior distribution using Monte-Carlo REINFORCE~\cite{williams1992simple} algorithm. During testing, we use a beam of 5 to approximate the whole search space for path finder. We follow MINERVA~\cite{das2017go} to set the maximum reasoning length to 3, which lowers the burden for the path-reasoner model. For both datasets, we set the embedding size $E$ to 200, the history embedding size $H$ to 200, the convolution kernel feature size $D$ to 128, we set the hidden size of MLP for both path finder and path reasoner to 400.
\begin{table}[t]
\centering
\small
\begin{tabular}{|l|c|c|c|c|}
\hline
Dataset & \#Ent & \#R & \#Triples & \#Tasks\\
\hline
FB15k-237 & 14,505 & 237 & 310,116 & 20\\
\hline
NELL-995 & 75,492 & 200 & 154,213 & 12\\
\hline
\end{tabular}
\caption{Dataset statistics.}
\label{tab:stat}
\end{table}
\subsection{Quantitative Results}
We mainly compare with the embedding-based algorithms~\cite{bordes2013translating,lin2015learning,ji2015knowledge,wang2014knowledge}, PRA~\cite{lao2011random}, MINERVA~\cite{das2017go}, DeepPath~\cite{xiong2017deeppath} and Chain-of-Reasoning~\cite{das2016chains}, besides, we also take our standalone CNN path-reasoner from \textsc{Diva}. Besides, we also try to directly maximize the marginal likelihood $p(r|e_s, e_d) = \sum_L p(L|e_s, e_d) p(r|L)$ using only the prior and likelihood model following MML~\cite{guu2017language}, which enables us to understand the superiority of introducing an approximate posterior. Here we first report our results for NELL-995 in~\autoref{tab:nell-result}, which is known to be a simple dataset and many existing algorithms already approach very significant accuracy. Then we test our methods in FB15k~\cite{toutanova2015representing} and report our results in~\autoref{tab:fb-result}, which is much harder than NELL and arguably more relevant for real-world scenarios.
\begin{table}[t]
\centering
\small
\begin{tabular}{|l|c|c|}
\hline
Model & 12-rel MAP & 9-rel MAP\\
\hline
RPA~\cite{lao2011random} & 67.5 & - \\
\hline
TransE~\cite{bordes2013translating} & 75.0 &- \\
\hline
TransR~\cite{lin2015learning} & 74.0 & -\\
\hline
TransD~\cite{ji2015knowledge} & 77.3 & -\\
\hline
TransH~\cite{wang2014knowledge} & 75.1 & -\\
\hline
MINERVA~\cite{das2017go} & - &     \textbf{88.2}\\
\hline
DeepPath~\cite{xiong2017deeppath} & 79.6 &    80.2\\
\hline
RNN-Chain~\cite{das2016chains}     & 79.0 & 80.2 \\
\hline
\hline
CNN Path-Reasoner    & 82.0 & 82.2 \\
\hline
\textsc{Diva} & \textbf{88.6} & 87.9 \\
\hline
\end{tabular}
\caption{MAP results on the NELL dataset. Since MINERVA~\cite{das2017go} only takes 9 relations out of the original 12 relations, we report the known results for both version of NELL-995 dataset.}
\label{tab:nell-result}
\end{table}

\begin{table}[t]
\centering
\small
\begin{tabular}{|l|c|}
\hline
Model & 20-rel MAP\\
\hline
PRA~\cite{lao2011random} & 54.1\\
\hline
TransE~\cite{bordes2013translating} & 53.2 \\
\hline
TransR~\cite{lin2015learning} & 54.0 \\
\hline
MINERVA~\cite{das2017go} & 55.2 \\
\hline
DeepPath~\cite{xiong2017deeppath} & 57.2 \\
\hline
RNN-Chain~\cite{das2016chains} & 51.2 \\
\hline
\hline
CNN Path-Reasoner & 54.2 \\
\hline
MML~\cite{guu2017language} & 58.7 \\
\hline
\textsc{Diva} & \textbf{59.8} \\
\hline
\end{tabular}
\caption{Results on the FB15k dataset, please note that MINERVA's result is obtained based on our own implementation.}
\label{tab:fb-result}
\end{table}
Besides, we also evaluate our model on FB-15k 20-relation subset with HITS@N score. Since our model only deals with the relation classification problem $(e_s, ?, e_d)$ with $e_d$ as input, so it's hard for us to directly compare with MINERVA~\cite{das2017go}. However, here we compare with chain-RNN~\cite{das2016chains} and CNN Path-Reasoner model, the results are demonstrated as~\autoref{tab:fb-result-hit}. Please note that the HITS@N score is computed against relation rather than entity.
\begin{table}[!htb]
\small
\centering
\begin{tabular}{|l|c|c|}
\hline
Model & HITS@3 & HITS@5\\
\hline
RNN-Chain~\cite{das2016chains} & 0.80 & 0.82\\
\hline
CNN Path-Reasoner & 0.82 & 0.83 \\
\hline
\textsc{Diva} & \textbf{0.84} & \textbf{0.86}\\
\hline
\end{tabular}
\caption{HITS@N results on the FB15k dataset}
\label{tab:fb-result-hit}
\end{table}

\paragraph{Result Analysis}
We can observe from the above tables~\autoref{tab:fb-result} and~\autoref{tab:nell-result} that our algorithm has significantly outperformed most of the existing algorithms and achieves a very similar result as MINERVA~\cite{das2017go} on NELL dataset and achieves state-of-the-art results on FB15k. We conclude that our method is able to deal with more complex reasoning scenarios and is more robust to the adversarial examples. Besides, we also observe that our CNN Path-Reasoner can outperform the RNN-Chain~\cite{das2016chains} on both datasets, we speculate that it is due to the short lengths of reasoning chains, which can extract more useful information from the reasoning chain. 

\begin{figure*}[t]
\centering
\includegraphics[width=1.0\linewidth]{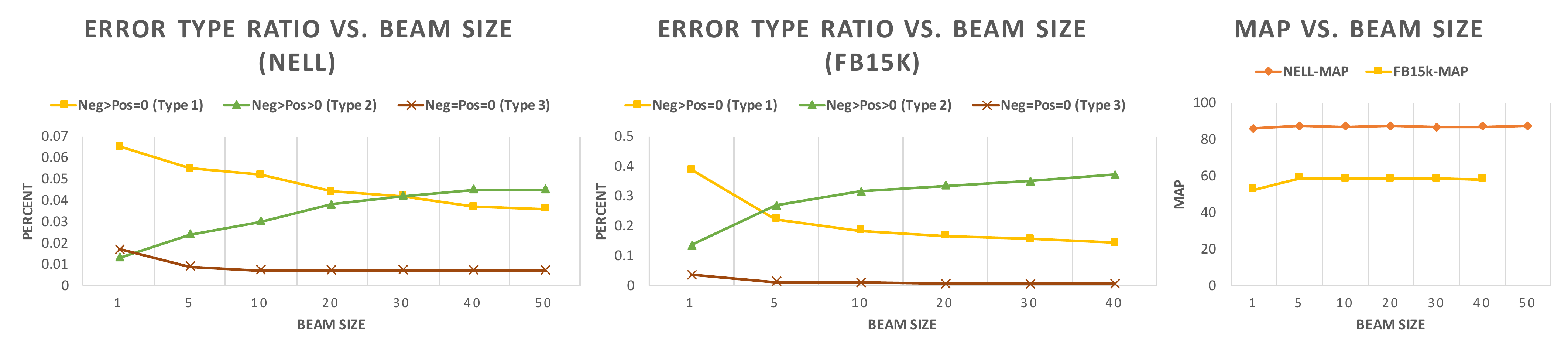}
\caption{MAP results varying beam size and the error type's occurrence w.r.t to beam size. A beam size that is too large or too small would cause performance to drop.}
\label{fig:beam-vis}
\end{figure*}
\begin{figure}[t]
\centering
\includegraphics[width=1.0\linewidth]{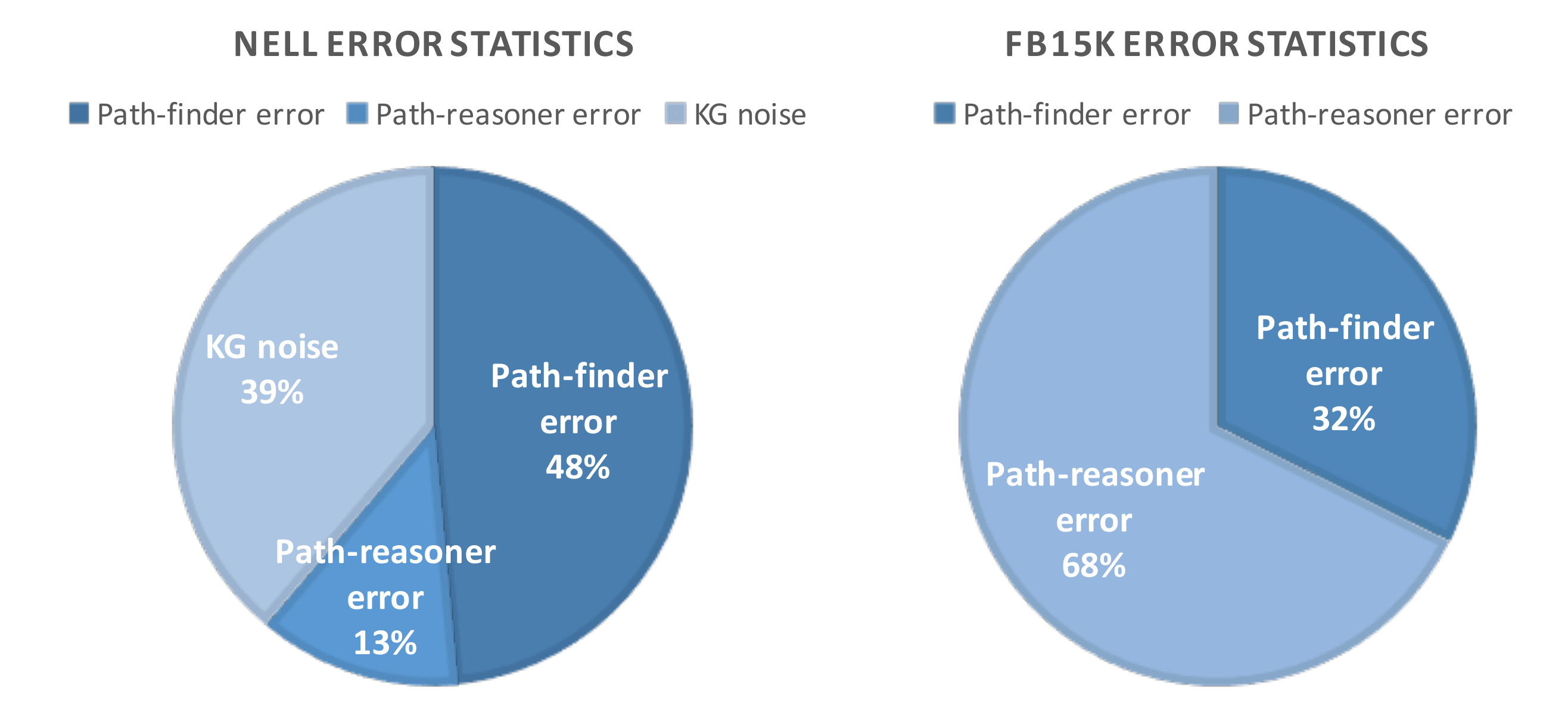}
\caption{Error analysis for the NELL and FB15k link prediction task. Since FB15k dataset uses placeholders for entities, we are not able to analyze whether the error comes from KG noise.} 
\label{fig:error-analysis}
\end{figure}
From these two pie charts in~\autoref{fig:error-analysis}, we can observe that in NELL-995, very few errors are coming from the path reasoner since the path length is very small. A large proportion only contains a single hop. In contrast, most of the failures in the FB15k dataset are coming from the path reasoner, which fails to classify the multi-hop chain into correct relation. This analysis demonstrates that FB15k is much harder dataset and may be closer to real-life scenarios.

\begin{table*}[t]
\centering
\small
\begin{tabular}{lll}\toprule
\multicolumn{1}{l}{Type} & \multicolumn{1}{l}{Reasoning Path} & \multicolumn{1}{l}{Score} \\ \midrule
Negative & athleteDirkNowitzki $\rightarrow$  \textit{(athleteLedSportsteam)} $\rightarrow$  sportsteamMavericks & 0.98 \\
Positive  & athleteDirkNowitzki $\rightarrow$ \textit{(athleteLedSportsteam)} $\rightarrow$ sportsteamDallasMavericks & 0.96 \\
\textit{Explanation} & \textit{``maverick'' is equivalent to ``dallas-maverick'', but treated as negative sample} & - \\ \midrule
Negative & athleteRichHill $\rightarrow$ \textit{(personBelongsToOrganization)} $\rightarrow$ sportsteamChicagoCubs & 0.88 \\
Positive & athleteRichHill $\rightarrow$ \textit{(personBelongsToOrganization)} $\rightarrow$ sportsteamBlackhawks & 0.74 \\
\textit{Explanation} & \textit{Rich Hill plays in both sportsteam but the knowledge graph only include one} & - \\
\midrule
\multirow{2}*{Negative} & coachNikolaiZherdev $\rightarrow$ \textit{(athleteHomeStadium)} $\rightarrow$ stadiumOreventvenueGiantsStadium & \\
&  $\rightarrow$   \textit{(teamHomestadium$^{-1}$)} $\rightarrow$  sportsteamNewyorkGiants & 0.98 \\
\multirow{2}*{Positive} & coachNikolaiZherdev $\rightarrow$ \textit{(athleteHomeStadium)} $\rightarrow$ stadiumOreventvenueGiantsStadium & \\
& $\rightarrow$  \textit{(teamHomestadium$^{-1}$)} $\rightarrow$  sportsteam-rangers & 0.72 \\
\textit{Explanation} &\textit{The home stadium accommodates multiple teams, therefore the logic chain is not valid } & - \\
\bottomrule
\end{tabular}
\caption{The three samples separately indicates three frequent error types, the first one belongs to ``duplicate entity'', the second one belongs to ``missing entity'', while the last one is due to ``wrong reasoning''. Please note that the parenthesis terms denote relations while the non-parenthesis terms denote entities.}
\label{tab:failure}
\end{table*}

\subsection{Beam Size Trade-offs}
Here we are especially interested in studying the impact of different beam sizes in the link prediction tasks. With larger beam size, the path finder can obtain more linking paths, meanwhile, more noises are introduced to pose greater challenges for the path reasoner to infer the relation. With smaller beam size, the path finder will struggle to find connecting paths between positive entity pairs, meanwhile eliminating many noisy links. Therefore, we first mainly summarize three different types and investigate their changing curve under different beam size conditions: 
\begin{enumerate}
\item No paths are found for positive samples, while paths are found for negative samples, which we denote as Neg$>$Pos=0.
\item Both positive samples and negative samples found paths, but the reasoner assigns higher scores to negative samples, which we denote as Neg$>$Pos$>$0.
\item Both negative and positive samples are not able to find paths in the knowledge graph, which we denote as Neg=Pos=0. 
\end{enumerate}
We draw the curves for MAP and error ratios in~\autoref{fig:beam-vis} and we can easily observe the trade-offs, we found that using beam size of 5 can balance the burden of path-finder and path-reasoner optimally, therefore we keep to this beam size for the all the experiments.

\subsection{Error Analysis}
In order to investigate the bottleneck of \textsc{Diva}, we take a subset from validation dataset to summarize the causes of different kinds of errors. Roughly, we classify errors into three categories, 1) KG noise: This error is caused by the KG itself, e.g some important relations are missing; some entities are duplicate; some nodes do not have valid outgoing edges. 2) Path-Finder error: This error is caused by the path finder, which fails to arrive destination. 3) Path-Reasoner error: This error is caused by the path reasoner to assign a higher score to negative paths. Here we draw two pie charts to demonstrate the sources of reasoning errors in two reasoning tasks.

\subsection{Failure Examples}
We also show some failure samples in~\autoref{tab:failure} to help understand where the errors are coming from. We can conclude that the ``duplicate entity'' and ``missing entity'' problems are mainly caused by the knowledge graph or the dataset, and the link prediction model has limited capability to resolve that. In contrast, the ``wrong reasoning'' problem is mainly caused by the reasoning model itself and can be improved with better algorithms.

\section{Conclusion}
\label{sec:conclusion}
In this paper, we propose a novel variational inference framework for knowledge graph reasoning. In contrast to prior studies that use a random walk with restarts~\cite{lao2011random} and explicit reinforcement learning path finding~\cite{xiong2017deeppath}, we situate our study in the context of variational inference in latent variable probabilistic graphical models. Our framework seamlessly integrates the path-finding and path-reasoning processes in a unified probabilistic framework, leveraging the strength of neural network based representation learning methods. Empirically, we show that our method has achieved the state-of-the-art performances on two popular datasets.

\section{Acknowledgement}
The authors would like to thank the anonymous reviewers for their thoughtful comments. This research
was sponsored in part by the Army Research Laboratory under cooperative agreements W911NF09-2-0053 and NSF IIS 1528175. The views and conclusions contained herein are those of the authors and should not be interpreted as representing the official policies, either expressed or implied, of the Army Research Laboratory or the U.S. Government. The U.S. Government is authorized to reproduce and distribute reprints for Government purposes notwithstanding any copyright notice herein.

\bibliography{naaclhlt2018}
\bibliographystyle{acl_natbib}
\appendix


\end{document}